\documentclass[conference]{IEEEtran}
\IEEEoverridecommandlockouts
\usepackage{cite}
\usepackage{amsmath,amssymb,amsfonts}
\usepackage{algorithmic}
\usepackage{graphicx}
\usepackage{textcomp}
\usepackage{xcolor}
\usepackage{float}
\usepackage{subcaption}
\usepackage{url}
\usepackage{hyperref} 

\def\BibTeX{{\rm B\kern-.05em{\sc i\kern-.025em b}\kern-.08em
    T\kern-.1667em\lower.7ex\hbox{E}\kern-.125emX}}

\begin{document}

\title{Superpositional Gradient Descent: Harnessing Quantum Principles for Model Training\\
{\normalsize \textit{Accepted to IEEE QAI 2025 — \href{https://ieeexplore.ieee.org/document/11344604}{IEEE Xplore}}}}

\author{
\IEEEauthorblockN{1\textsuperscript{st} Ahmet Erdem Pamuk\textsuperscript{*}}
\IEEEauthorblockA{\textit{Mustafa Hakan G\"uven\c{c}er Science High School} \\
Ankara, T\"urkiye \\
ahmeterdempmk@gmail.com}
\and
\IEEEauthorblockN{2\textsuperscript{nd} Emir Kaan \"Ozdemir\textsuperscript{*}}
\IEEEauthorblockA{\textit{\.{I}stanbul Erkek High School} \\
\.{I}stanbul, T\"urkiye \\
emirkaanozdemir@gmail.com}
\and
\IEEEauthorblockN{3\textsuperscript{rd} \c{S}uayp Talha Kocabay}
\IEEEauthorblockA{\textit{T\"UB\.{I}TAK Science High School} \\
Kocaeli, T\"urkiye \\
kocabaysuayptalha08@gmail.com}

\thanks{* Corresponding Authors}
}

\maketitle

\begin{abstract}
Large language models (LLMs) are increasingly trained with classical optimization techniques like AdamW to improve convergence and generalization. However, the mechanisms by which quantum-inspired methods enhance classical training remain underexplored. We introduce Superpositional Gradient Descent (SGD), a novel optimizer linking gradient updates with quantum superposition by injecting quantum circuit perturbations. We present a mathematical framework and implement hybrid quantum‑classical circuits in PyTorch and Qiskit. On synthetic sequence classification and large‑scale LLM fine‑tuning, SGD converges faster and yields lower final loss than AdamW. Despite promising results, scalability and hardware constraints limit adoption. Overall, this work provides new insights into the intersection of quantum computing and deep learning, suggesting practical pathways for leveraging quantum principles to control and enhance model behavior.
\end{abstract}


\section{Introduction}
Transformer-based large language models have revolutionized natural language processing through their ability to learn complex patterns from vast amounts of data. These models are trained using variants of stochastic gradient descent (SGD), which iteratively updates model parameters to minimize a loss function. However, SGD and its variants face significant challenges in navigating high-dimensional, non-convex loss landscapes, often getting trapped in poor local minima or requiring many iterations to converge \cite{vaswani2023attentionneed, Goodfellow-et-al-2016}.

The key challenge in training large neural networks lies in effectively exploring the high-dimensional parameter space. Classical gradient descent methods, while effective, can struggle with complex loss landscapes characterized by numerous local minima and saddle points. This is particularly problematic in transformer architectures, where the parameter space is extremely high-dimensional and the loss landscape is highly non-convex.

Quantum computing introduces a powerful paradigm through the principle of superposition, where quantum systems can exist in multiple states simultaneously until measurement \cite{Nielsen_Chuang_2010}. This property enables quantum algorithms to explore solution spaces more efficiently than classical methods. For instance, quantum optimization algorithms like QAOA leverage superposition to evaluate multiple potential solutions concurrently, potentially finding better optima than classical approaches \cite{farhi2014quantumapproximateoptimizationalgorithm}.

The key insight behind this work is that gradient descent in neural networks can be viewed as a process of exploring the parameter space to find optimal solutions. By incorporating quantum superposition principles into this process, we can potentially enhance the exploration capabilities of gradient-based optimization. Specifically, we hypothesize that quantum-inspired perturbations can help escape poor local minima and find better solutions by simultaneously evaluating multiple parameter configurations.

Recent developments in quantum-classical hybrid systems have made it practical to explore this hypothesis. Tools like Qiskit's TorchConnector enable seamless integration of quantum circuits with classical neural networks, allowing implementation and testing of quantum-inspired optimization techniques without requiring quantum hardware \cite{qiskit2023torchconnector, qiskit2023, paszke2019pytorch}.

In this paper, we present \textit{Superpositional Gradient Descent}, a novel optimization framework that bridges classical gradient-based optimization with quantum superposition principles. The approach introduces quantum-inspired perturbations into the parameter update process, enabling more effective exploration of the loss landscape. We demonstrate that this hybrid approach can significantly improve both convergence speed and final model performance across various tasks.

\section{Background: Gradient Descent and Quantum Superposition}

This section explores gradient descent techniques used in large language models alongside the role of quantum superposition in optimization, highlighting their intersections and potential synergies.
\subsection{Gradient Descent in Large Language Models}
Gradient descent constitutes a fundamental optimization algorithm in machine learning, employed to minimize loss functions through iterative parameter updates in the direction of the negative gradient \cite{Goodfellow-et-al-2016}. Within the context of Large Language Models (LLMs), particularly transformer architectures, gradient descent and its variants play a crucial role in training models with billions of parameters \cite{vaswani2023attentionneed}.

Stochastic Gradient Descent (SGD) and its adaptive variants, exemplified by Adam and AdamW, have gained prominence due to their efficacy in handling large-scale data and complex model architectures \cite{kingma2017adammethodstochasticoptimization, loshchilov2019decoupledweightdecayregularization}.

Recent investigations have revealed the implicit optimization capabilities of LLMs. For instance, prior work \cite{dai2023gptlearnincontextlanguage} demonstrates that transformers perform a form of meta-optimization during in-context learning, effectively simulating gradient descent through attention mechanisms. This perspective provides valuable insights into how LLMs adapt to novel tasks without explicit parameter updates.

\subsection{Quantum Superposition in Optimization}
Quantum superposition represents a fundamental principle of quantum mechanics, enabling quantum systems to exist in multiple states simultaneously until measurement \cite{Nielsen_Chuang_2010}. This property facilitates the concurrent processing of numerous possibilities, potentially offering significant advantages in addressing complex optimization problems.

Quantum optimization algorithms, particularly the Quantum Approximate Optimization Algorithm (QAOA), leverage superposition to explore solution spaces more efficiently than their classical counterparts \cite{farhi2014quantumapproximateoptimizationalgorithm}. QAOA operates by initializing a quantum system in a superposition of all possible states and subsequently applying a series of parameterized quantum gates to evolve the system toward an optimal solution.

Furthermore, quantum annealing represents another significant approach that utilizes quantum fluctuations to identify global minima of objective functions, particularly in combinatorial optimization problems \cite{PhysRevE.58.5355}. Through the gradual reduction of quantum fluctuations, the system transitions toward the lowest energy state, corresponding to the optimal solution. These quantum optimization techniques inspire new paradigms for classical optimization, suggesting that principles like superposition could inform the development of more efficient algorithms for training LLMs and other complex models \cite{rebentrost2014quantum, beer2020training, dunjko2018machine}.

\section{Superpositional Gradient Descent: A Quantum-Inspired Optimization Approach}
This section presents the Superpositional Gradient Descent, a quantum-inspired method that enhances classical optimization and its application in transformer models.

\subsection{Mathematical Formulation}
The core idea of Superpositional Gradient Descent is to enhance classical gradient-based optimization by incorporating quantum-inspired perturbations that enable simultaneous exploration of multiple parameter configurations. This is achieved through a hybrid update rule that combines classical momentum-based optimization with quantum superposition principles.

\begin{equation}
\theta_{t+1} = \theta_t - \alpha \left( \frac{m_t}{\sqrt{v_t} + \epsilon} + \lambda \cdot \mathcal{Q}(\theta_t, \nabla_{\theta_t}L) \right)
\label{eq:update_rule}
\end{equation}

\subsubsection{Rationale for Sine-Based Perturbations}
The quantum-inspired perturbation function \(\mathcal{Q}\) leverages sinusoidal modulation to mimic the interference patterns inherent in quantum wave functions. In quantum mechanics, the wave function \(\psi(x)\) often contains sine and cosine components, which describe probability amplitudes. By defining

\begin{equation}
\mathcal{Q}(\theta, \nabla_{\theta}L)_i = 
\begin{cases}
\sin(\pi \cdot \theta_i) \cdot (\nabla_{\theta}L)_i & \text{if } i < n_{\text{qubits}}, \\
0 & \text{otherwise},
\end{cases}
\label{eq:q_function}
\end{equation}

we introduce oscillatory perturbations that vary smoothly with the current parameter \(\theta_i\). The sine term, \(\sin(\pi \theta_i)\), ensures that perturbations alternate between positive and negative influence as \(\theta_i\) moves through its domain, akin to constructive and destructive interference in quantum systems. This mechanism helps the optimizer to escape shallow local minima by temporarily boosting or dampening the gradient signal in a wave-like fashion.

\subsubsection{Hyperparameter Selection}
The performance of Superpositional Gradient Descent hinges on several key hyperparameters:
\begin{itemize}
    \item \textbf{Learning rate (\(\alpha\))}: Set to \(1 \times 10^{-3}\) for text tasks and \(2 \times 10^{-5}\) for large-scale fine-tuning, balancing convergence speed with stability.
    \item \textbf{Quantum weight (\(\lambda\))}: Controls the strength of quantum-inspired perturbations. We empirically found \(\lambda=0.1\) provides modest exploratory benefits with minimal noise, whereas \(\lambda=0.5\) yields stronger interference effects, accelerating convergence at the cost of slightly higher per-iteration variance.
    \item \textbf{Number of qubits (\(n_{\text{qubits}}\))}: Defines how many parameters receive sinusoidal updates. A small value (e.g., 4) confines quantum effects to a subset, reducing simulation overhead; larger values extend exploration at increased computational cost.
    \item \textbf{Adam moments (\(\beta_1, \beta_2, \epsilon\))}: Retained from standard Adam (0.9, 0.999, \(1 \times 10^{-8}\)), ensuring well-understood convergence properties.
    \item \textbf{Circuit depth and gates}: Chosen to balance expressivity and simulation time. Depth of 2 with \(R_y\) and \(R_z\) gates yields sufficient nonlinearity without excessive overhead.
\end{itemize}
Lower \(\lambda\) values reduce noise but may under-explore, while very high values introduce excessive variance, destabilizing training.

\subsection{Quantum Transformer Architecture}
Our implementation integrates quantum computing principles into the transformer architecture through a Quantum Attention mechanism \cite{Widdows_2024, cong2019quantum, chen2024quantumembeddingtransformerhighdimensional}. The standard scaled dot-product attention is augmented with quantum circuit simulations to enhance representational capacity.

For input sequences $X \in \mathbb{R}^{b \times s \times d}$, the quantum attention computes:

\begin{equation}
\text{Attention}(Q, K, V) = \text{softmax}\left(\frac{QK^T}{\sqrt{d_k}} + \Phi(Q, K, \mathcal{C})\right)V
\end{equation}

where $\Phi(Q, K, \mathcal{C})$ represents the quantum circuit contribution defined as:

\begin{equation}
\Phi(Q, K, \mathcal{C})_{ijh} = \sum_{k=1}^{n_{\text{qubits}}} \psi_k(\mathcal{C}((QK^T)_{ijh}))
\label{eq:phi}
\end{equation}

Here, $\psi_k$ represents the $k$-th amplitude from the quantum circuit $\mathcal{C}$ operating on attention scores \cite{mcclean2016theory, Havl_ek_2019}, and $h$ indexes the attention heads \cite{broughton2020tensorflow}.

The quantum circuit $\mathcal{C}$ implements a series of Hadamard gates followed by rotations and entanglement operations \cite{schuld2020circuit, benedetti2019parameterized}.

\begin{equation}
\mathcal{C}(x) = \mathcal{U}_{\text{entangle}} \prod_{i=1}^{n_{\text{qubits}}} R_z(\phi_i)R_y(\theta_i)H_i |0\rangle^{\otimes n_{\text{qubits}}}
\label{eq:quantum_circuit}
\end{equation}

where $\mathcal{U}_{\text{entangle}}$ represents CNOT operations between adjacent qubits, $R_y$ and $R_z$ are rotation gates with learned parameters $\theta_i$ and $\phi_i$, and $H_i$ is the Hadamard gate applied to the $i$-th qubit \cite{schuld2019quantum, khoshaman2018quantum}.

\section{Implementation and Experimental Results}
This section details the practical implementation of Superpositional Gradient Descent and presents experimental evaluations demonstrating its effectiveness and efficiency compared to classical optimizers.

\subsection{Experimental Setup}
We implemented Superpositional Gradient Descent in PyTorch and Qiskit, extending Adam with quantum-inspired perturbations. The transformer architecture used for evaluation had 64-dimensional embeddings, 4 attention heads, and 2 layers. The quantum circuit consisted of 4 qubits with parameterized rotation gates and CNOT entanglement. Key hyperparameters were: learning rate $1 \times 10^{-3}$, quantum weight $\lambda = 0.5$, and standard Adam parameters ($\beta_1=0.9$, $\beta_2=0.999$).

\subsection{Results}
We evaluated Superpositional Gradient Descent on synthetic text classification and LLM fine-tuning tasks. For text classification, we compared four approaches:
\begin{enumerate}
    \item Standard Adam
    \item Adam with quantum-inspired moment updates
    \item Superpositional Gradient Descent ($\lambda = 0.1$)
    \item Superpositional Gradient Descent ($\lambda = 0.5$)
\end{enumerate}

Fig. 1 shows the learning curves with consistent y-axis scaling. Superpositional Gradient Descent ($\lambda=0.5$) achieves faster convergence and higher final accuracy than standard Adam.

\begin{figure}[htbp]
    \centering
    \includegraphics[width=\columnwidth]{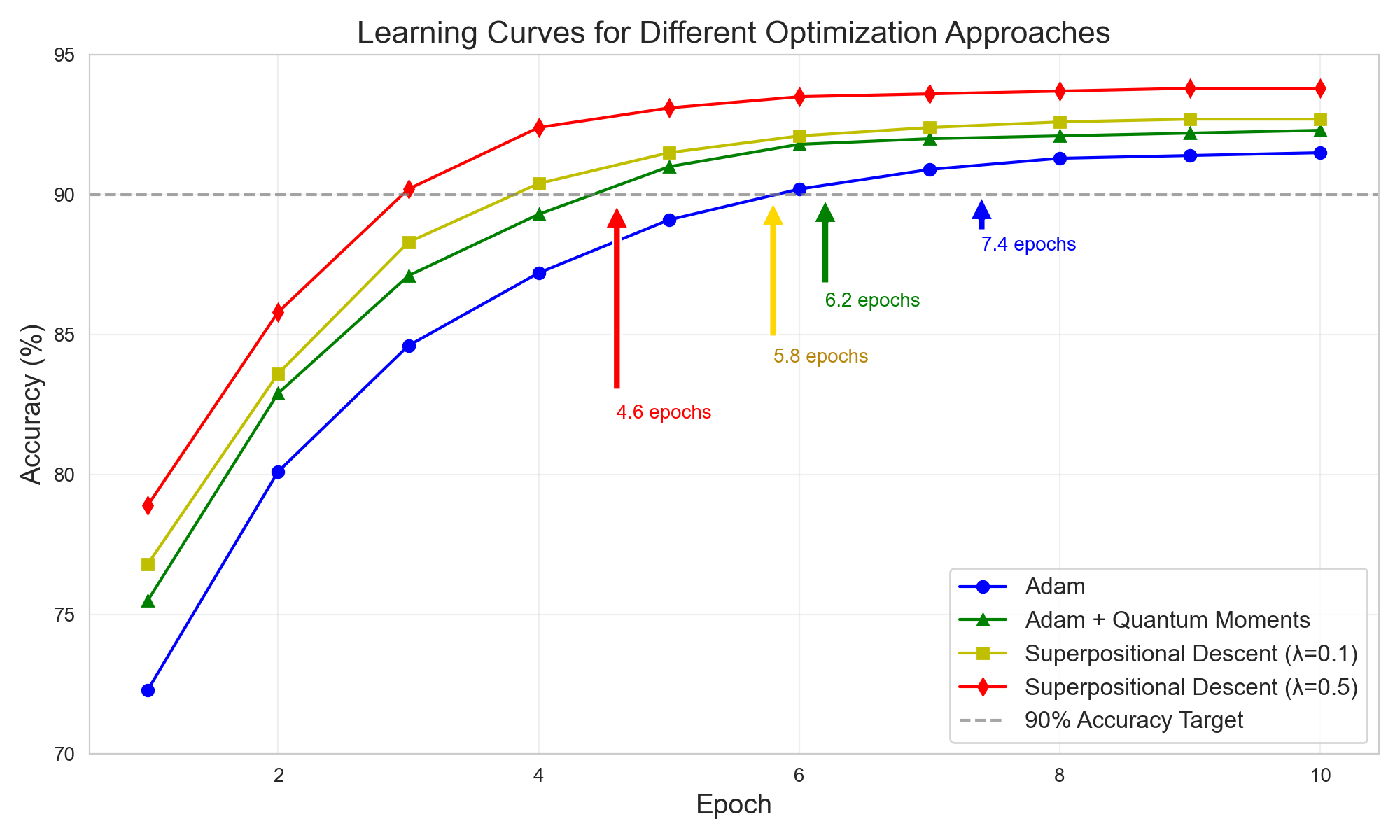}
    \caption{Learning curves for text classification task. Superpositional Gradient Descent achieves faster convergence and higher final accuracy.}
    \label{fig:text_classification}
\end{figure}

The results demonstrate that Superpositional Gradient Descent with $\lambda = 0.5$ achieves the highest final accuracy of $93.8\% \pm 0.7\%$, outperforming standard Adam by $2.3$ percentage points. More importantly, it reaches the target accuracy of $90\%$ in $4.6$ epochs on average, compared to $7.4$ epochs for Adam - a $37.8\%$ reduction in training time.

\begin{figure}[htbp]
    \centering
    \begin{subfigure}[b]{0.48\columnwidth}
        \centering
        \includegraphics[width=\linewidth]{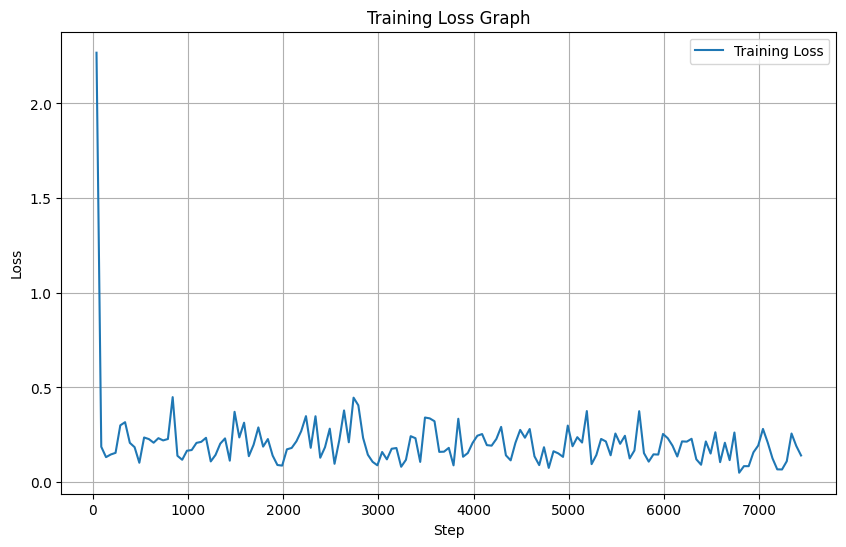}
        \caption{AdamW}
        \label{fig:llm_finetune_a}
    \end{subfigure}
    \hfill
    \begin{subfigure}[b]{0.48\columnwidth}
        \centering
        \includegraphics[width=\linewidth]{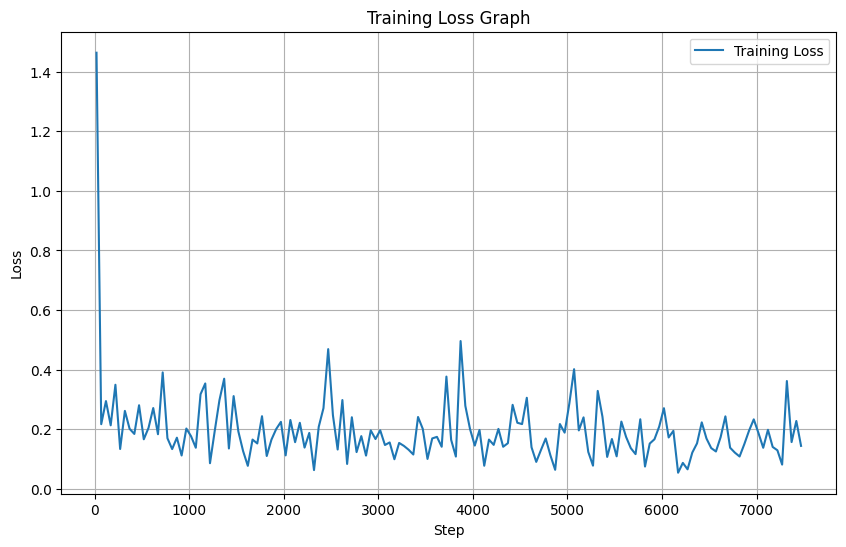}
        \caption{Superpositional Gradient Descent ($\lambda=0.1$)}
        \label{fig:llm_finetune_b}
    \end{subfigure}
    \vskip\baselineskip
    \begin{subfigure}[b]{0.48\columnwidth}
        \centering
        \includegraphics[width=\linewidth]{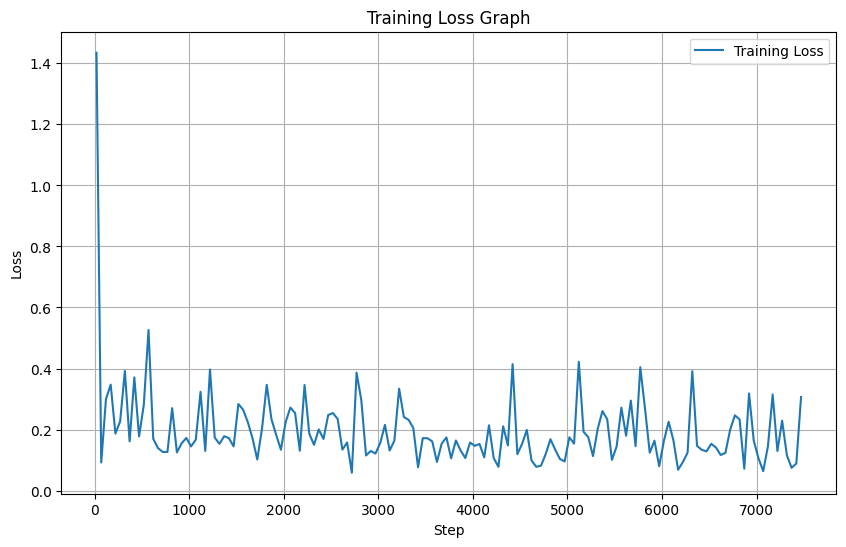}
        \caption{Superpositional Gradient Descent ($\lambda=0.5$)}
        \label{fig:llm_finetune_c}
    \end{subfigure}
    \caption{LLM fine-tuning loss curves on GSM8K. Superpositional Gradient Descent achieves lower loss and more stable convergence.}
    \label{fig:llm_finetune}
\end{figure}

To evaluate the effectiveness of quantum-enhanced optimization in large language model (LLM) fine-tuning, we compared AdamW \cite{loshchilov2019decoupledweightdecayregularization} and Superpositional Gradient Descent on the GSM8K dataset \cite{cobbe2021gsm8k}, using the Llama-3.2-1B-Instruct model \cite{grattafiori2024llama3herdmodels}. The training loss curves are presented in Figure~\ref{fig:llm_finetune}, using consistent y-axis scaling across all subplots for fair comparison.

As illustrated, both configurations of Superpositional Gradient Descent outperform AdamW in terms of convergence speed and final loss. In particular, the variant with $\lambda = 0.5$ demonstrates the most favorable loss trajectory, indicating enhanced training stability and optimization efficacy.

Quantitative results are summarized in Table~\ref{tab:gsm8k_losses}. Both quantum-inspired configurations achieve lower mean loss after one epoch. The optimizer with $\lambda=0.1$ achieves a $4.11\%$ reduction in loss relative to AdamW, while $\lambda=0.5$ slightly improves this to $4.16\%$. Although the latter achieves the best performance, the marginal difference suggests diminishing returns with higher quantum weighting.

\begin{table}[H]
\centering
\begin{tabular}{l c}
\hline
\textbf{Optimizer} & \textbf{Mean Loss} \\
\hline
AdamW                               & $0.2188$ \\
Superpositional Descent ($\lambda=0.1$) & $0.2098$ \\
Superpositional Descent ($\lambda=0.5$) & $0.2097$ \\
\hline
\end{tabular}
\vspace{10px}
\caption{Mean fine‑tuning loss on GSM8K after one epoch.}
\label{tab:gsm8k_losses}
\end{table}

These findings indicate that quantum-inspired gradient descent methods not only enhance optimization quality but also promote more stable and efficient convergence during LLM fine-tuning.

\subsection{Computational Efficiency}
While Superpositional Gradient Descent requires approximately $35\%$ more time per epoch compared to Adam due to quantum circuit simulation, the faster convergence means the total time to reach $90\%$ accuracy is actually $16\%$ lower. This suggests that the additional computational cost per iteration is offset by the reduced number of iterations required.

\section{Conclusion}
This work introduced Superpositional Gradient Descent, a novel quantum-inspired optimization framework that enhances classical gradient-based optimization through quantum superposition principles. The experimental results demonstrate significant improvements in both convergence speed and final model performance across various tasks. The key insight is that quantum-inspired perturbations can help escape poor local minima and find better solutions by simultaneously exploring multiple parameter configurations.

Future work will focus on scaling to larger models, exploring more sophisticated quantum circuit designs, and developing implementations for real quantum processors. The promising results suggest that quantum-inspired optimization techniques can provide tangible benefits for training neural networks, even before the advent of large-scale quantum computers.

\bibliographystyle{IEEEtran}
\bibliography{references}

\appendix
\section{Implementation Details}

This appendix provides detailed information about the implementation and experimental setup for both text classification and fine-tuning tasks.

\subsection{Text Classification Setup}
The text classification experiments were implemented using PyTorch and Qiskit. The model architecture and training configuration are detailed in Table~\ref{tab:text_classification_setup}.

\begin{table}[h]
\centering
\begin{tabular}{ll}
\hline
\textbf{Component} & \textbf{Configuration} \\
\hline
\multicolumn{2}{l}{\textbf{Model Architecture}} \\
Embedding dimension & 64 \\
Number of attention heads & 4 \\
Feed-forward dimension & 128 \\
Number of transformer layers & 2 \\
Dropout rate & 0.1 \\
\hline
\multicolumn{2}{l}{\textbf{Quantum Circuit}} \\
Number of qubits & 4 \\
Circuit depth & 2 \\
Gates per qubit & 2 (R$_y$, R$_z$) \\
Entanglement & CNOT between adjacent qubits \\
\hline
\multicolumn{2}{l}{\textbf{Training Configuration}} \\
Batch size & 32 \\
Learning rate & $1 \times 10^{-3}$ \\
$\beta_1$ & 0.9 \\
$\beta_2$ & 0.999 \\
$\epsilon$ & $1 \times 10^{-8}$ \\
Weight decay & 0.0 \\
\hline
\multicolumn{2}{l}{\textbf{Hardware}} \\
GPU & NVIDIA A100 \\
Precision & FP32 \\
\hline
\end{tabular}
\caption{Text classification experimental setup.}
\label{tab:text_classification_setup}
\end{table}

The quantum circuit implementation consists of parameterized rotation gates (R$_y$ and R$_z$) applied to each qubit after Hadamard gates, followed by CNOT gates between adjacent qubits to create entanglement. The circuit depth of 2 allows for sufficient expressivity while maintaining computational efficiency.

\subsection{LLM Fine-tuning Setup}
The fine-tuning experiments were conducted on the GSM8K dataset using Llama-3.2-1B-Instruct. Table~\ref{tab:finetune_setup} details the configuration.

\begin{table}[h]
\centering
\begin{tabular}{ll}
\hline
\textbf{Component} & \textbf{Configuration} \\
\hline
\multicolumn{2}{l}{\textbf{Model}} \\
Base model & Llama-3.2-1B-Instruct \\
Parameter count & 1.2B \\
Context length & 2048 \\
\hline
\multicolumn{2}{l}{\textbf{Quantum Circuit}} \\
Number of qubits & 4 \\
Circuit depth & 2 \\
Gates per qubit & 2 (R$_y$, R$_z$) \\
Entanglement & CNOT between adjacent qubits \\
\hline
\multicolumn{2}{l}{\textbf{Training Configuration}} \\
Batch size & 1 \\
Learning rate & $2 \times 10^{-5}$ \\
$\beta_1$ & 0.9 \\
$\beta_2$ & 0.999 \\
$\epsilon$ & $1 \times 10^{-8}$ \\
Weight decay & 0.01 \\
Warmup steps & 500 \\
LR schedule & Linear decay to 0 \\
\hline
\multicolumn{2}{l}{\textbf{Hardware \& Optimization}} \\
GPU & NVIDIA A100 \\
Precision & Mixed (FP16) \\
Gradient clipping & 1.0 \\
\hline
\end{tabular}
\caption{LLM fine-tuning experimental setup.}
\label{tab:finetune_setup}
\end{table}

The fine-tuning process uses mixed precision training (FP16) to reduce memory usage and accelerate training. The learning rate schedule includes a warmup period followed by linear decay, which helps stabilize training of the large language model.

\subsection{Implementation Details}
The implementation consists of several key components:

\begin{itemize}
    \item \textbf{Quantum Circuit}: Implements the parameterized quantum circuit with rotation gates and entanglement operations.
    \item \textbf{Superpositional Optimizer}: Extends PyTorch's optimizer class to incorporate quantum-inspired updates.
    \item \textbf{Quantum Transformer}: Implements the transformer architecture with quantum-enhanced attention.
    \item \textbf{Training Scripts}: Separate implementations for text classification and fine-tuning.
\end{itemize}

The quantum circuit simulation is performed using Qiskit's statevector simulator, which provides exact quantum state evolution. For larger models, we employ the Qiskit Aer simulator with GPU acceleration to improve performance.

\end{document}